\documentclass[conference]{IEEEtran}
\IEEEoverridecommandlockouts

\usepackage{multirow}
\usepackage{amsmath}
\usepackage{amssymb}
\usepackage{graphicx}
\graphicspath{{./figures/}}

\usepackage{algorithm}
\usepackage{algorithmic}

\usepackage[caption=false,font=footnotesize]{subfig}
\usepackage{pifont}
\usepackage{multirow}
\usepackage{hyperref}

\usepackage{siunitx}
\usepackage{glossaries}
\newacronym{sota}{SOTA}{state-of-the-art}
\newacronym{lidar}{LiDAR}{LiDAR}
\newacronym{fov}{FOV}{Field-of-View}
\newacronym{v2x}{V2X}{Vehicle-to-Everything}
\newacronym{rtk}{RTK}{Real Time Kinematic}
\newacronym{fhd}{FHD}{Full High Definition}
\newacronym{fps}{FPS}{Frames Per Second}
\newacronym{ptp}{PTP}{Precision Time Protocol}
\newacronym{ntp}{NTP}{Network Time Protocol}
\newacronym{gps}{GPS}{Global Positioning System}
\newacronym{imu}{IMU}{Inertial Measurement Unit}
\newacronym{HD maps}{HD maps}{High-Definition maps}
\newacronym{mAP}{mAP}{mean Average Precision}

\title{\LARGE \bf DrivIng: A Large-Scale Multimodal Driving Dataset with Full Digital Twin Integration}

\begin{document}

\author{Dominik Rößle$^{1}$, Xujun Xie$^{1}$, Adithya Mohan$^{1}$, Venkatesh Thirugnana Sambandham$^{1}$,\\Daniel Cremers$^{2}$, Torsten Schön$^{1}$
\thanks{$^{1}$Dominik Rößle, Xujun Xie, Adithya Mohan, Venkatesh Thirugnana Sambandham, and Torsten Schön are with the Department of Computer Science and AImotion Bavaria, Technische Hochschule Ingolstadt, 85049 Ingolstadt, Germany
        {\tt\small \{dominik.roessle, xujun.xie, adithya.mohan, venkatesh.thirugnanasambandham, torsten.schoen\}@thi.de}}
\thanks{$^{2}$Daniel Cremers is with the School of Computation, Information and Technology, Technical University of Munich, 85748 Garching, Germany
        {\tt\small cremers@tum.de}}
\thanks{© 2026 IEEE. This is the author’s accepted manuscript of a paper accepted to the IEEE Intelligent Vehicles Symposium (IV), 2026.}
}

\maketitle

\begin{abstract}
    Perception is a cornerstone of autonomous driving, enabling vehicles to understand their surroundings and make safe, reliable decisions.
    Developing robust perception algorithms requires large-scale, high-quality datasets that cover diverse driving conditions and support thorough evaluation.
    Existing datasets often lack a high-fidelity digital twin, limiting systematic testing, edge-case simulation, sensor modification, and sim-to-real evaluations.
    To address this gap, we present DrivIng, a large-scale multimodal dataset with a complete geo-referenced digital twin of a $\mathbf{\sim\!18~km}$ route spanning urban, suburban, and highway segments.
    Our dataset provides continuous recordings from six RGB cameras, one LiDAR, and high-precision ADMA-based localization, captured across day, dusk, and night. All sequences are annotated at 10~Hz with 3D bounding boxes and track IDs across 12 classes, yielding $\mathbf{\sim\!1.2~\text{million}}$ annotated instances.
    Alongside the benefits of a digital twin, DrivIng enables a 1-to-1 transfer of real traffic into simulation, preserving agent interactions while enabling realistic and flexible scenario testing.
    To support reproducible research and robust validation, we benchmark DrivIng with state-of-the-art perception models and publicly release the dataset, digital twin, HD map, and codebase via \href{https://github.com/cvims/DrivIng}{\texttt{https://github.com/cvims/DrivIng}}.
\end{abstract}

\begin{figure*}
    \centering
    \includegraphics[width=0.99\linewidth, height=0.5\linewidth]{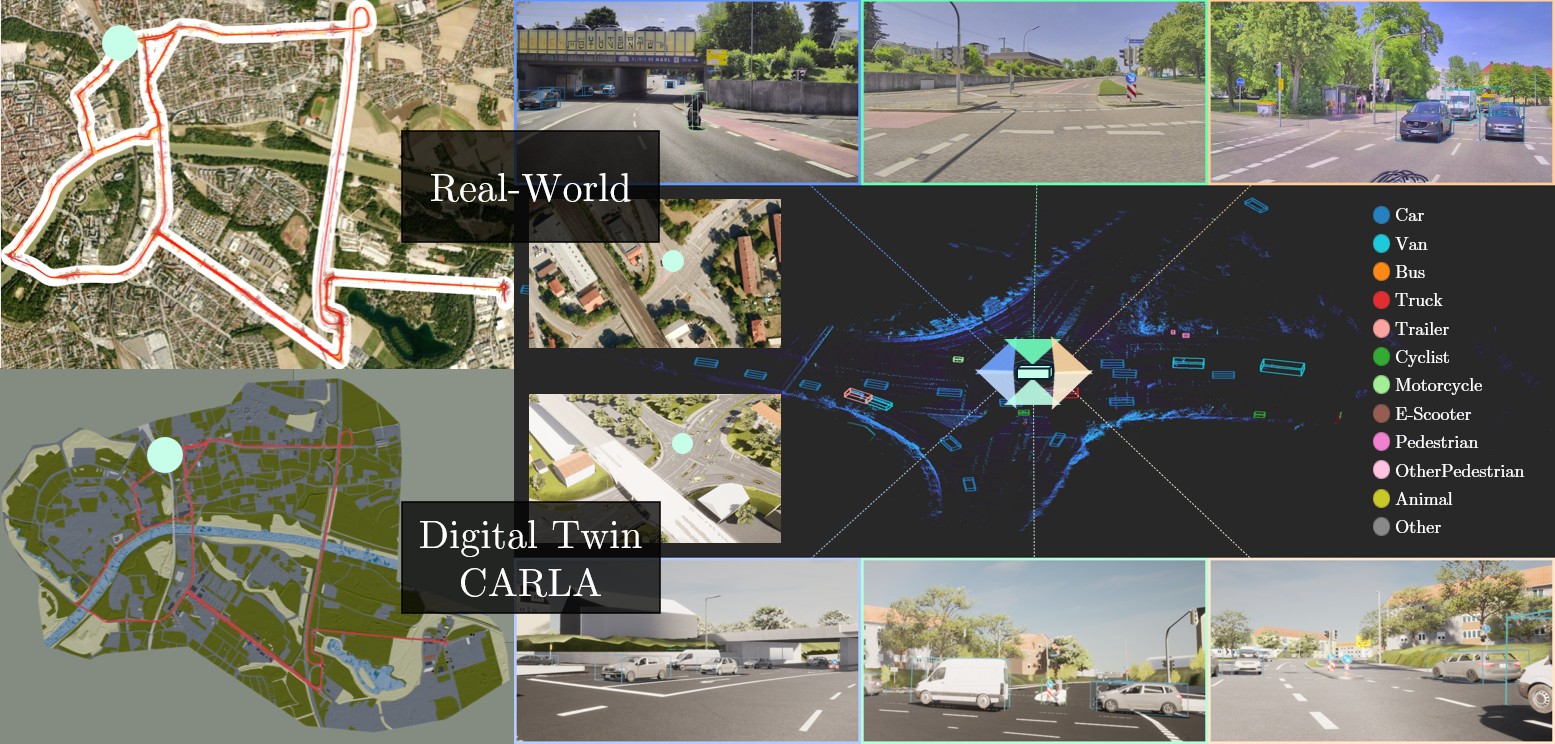}
    \caption{This visualization illustrates the core features of DrivIng and its digital twin. The left panel shows a real-world satellite view of the track and its fully geo-referenced digital twin, aligned with a location marker indicating the vehicle’s position. The right panel presents the synchronized sensor suite, including six camera views and a LiDAR frame. The top row displays real-world images, while the bottom row shows the corresponding CARLA simulation with all real-world objects precisely mapped. All images and the LiDAR frame include class-colored 3D bounding boxes for clear object distinction. Satellite image \textcopyright{} Esri, i-cubed, USDA, USGS, AEX, GeoEye, Getmapping, Aerogrid, IGN, IGP, UPR-EGP, and the GIS User Community.}
    \label{fig:teaser}
\end{figure*}

\section{Introduction}
Perception is fundamental to autonomous driving, delivering the essential understanding of a vehicle’s surroundings required for safe and reliable decision-making~\cite{TempCoBEV2024Roessle}.
Among perception tasks, robust object-level perception is essential and relies on large-scale, high-quality, precisely annotated data for accurate detection, tracking, and interpretation across diverse driving conditions~\cite{KITTI2013Geiger}.
Autonomous driving research involves perception tasks that must be executed reliably across various environments, including urban, suburban, and highway settings, each presenting unique challenges.
To improve robustness and accuracy, multi-sensor setups~\cite{PHP2022Roessle} equipped with precise geo-referencing systems are commonly used, providing richer environmental context and enhancing situational awareness~\cite{Urbaningv2x2025Roessle}.
Nevertheless, developing perception algorithms is challenging because it requires not only datasets that cover an immense range of real-world variations \cite{Waymo2020Sun, Zenseact2023alibeigi} but also robust validation to ensure model reliability.
Simulation environments provide a powerful solution to this limitation~\cite{AutonomousDrivingValidationDigitalTwins2024Pikner}.
They enable the modification of environmental conditions and the systematic evaluation~\cite{EnhRealFloatingCarObs2023Gerner, AIMotionChallengeResults2022Souza} of algorithms under edge cases~\cite{SimulationMatters2024Hu, AdvRobustnessDRL2025Mohan}.
Recent research in cooperative perception, where multiple agents share and fuse sensor data to mitigate occlusions and to improve the overall perception of surrounding objects~\cite{TempCoBEV2024Roessle, Urbaningv2x2025Roessle}, highlights the need for simulation-aided approaches that can replicate complex, synchronized multi-agent scenarios, which are often prohibitively expensive or logistically challenging to reproduce in the real world.
Despite the advantages of simulation, most existing large-scale driving datasets lack a comprehensive digital twin, limiting the ability to augment real-world data and rigorously benchmark perception algorithms.
To address this gap, we introduce DrivIng, a large-scale, multimodal driving dataset with full support for real-to-sim mapping.
By providing a digital twin of the recorded route, DrivIng bridges the domain gap and enables a wide range of applications, including systematic testing, the creation of complex multi-agent scenarios, and sim-to-real experiments that leverage both real-world and simulated data~\cite{NVIDIACosmos2025NVIDIA}.
The main contributions of \textbf{DrivIng} are:
\begin{enumerate}
    \item \textbf{Comprehensive real-world dataset:} Covers an approximately \SI{18}{\kilo\meter} route across urban, suburban, and highway environments, recorded with six RGB cameras offering 360$^\circ$ coverage and a roof-mounted LiDAR, under day, dusk, and night conditions.
    \item \textbf{High-frequency annotations:} Provided at 10~Hz with 3D bounding boxes for 12 object classes, yielding approximately 1.2 million labeled instances.
    \item \textbf{Fully-integrated data and validation testbed:} A digital twin of the entire recorded route enables simulation-based scenario replay, environmental modification, and systematic evaluation of perception algorithms.
    \item \textbf{Benchmark evaluations:} Conducted on real-world data using \gls{sota} camera and LiDAR perception models implemented in MMDetection3D~\cite{MMDet3d2020MMDet3dcontributors}.
    \item \textbf{Developer toolkit and public release:} Includes a nuScenes-format converter, dataset, codebase, the digital twin for reproducible research and a wide range of perception tasks with real-world and simulated data.
\end{enumerate}

\section{Related Work}
Large-scale datasets have been instrumental in advancing autonomous driving research, providing annotated data across a diverse range of environments for tasks such as 3D object detection, tracking, and motion forecasting.
Established datasets such as KITTI~\cite{KITTI2013Geiger}, nuScenes~\cite{NuScenes2020Caesar}, and Waymo Open Dataset~\cite{Waymo2020Sun} capture diverse real-world driving conditions and have become standard benchmarks for evaluating perception methods.
These datasets are typically structured as many short, independent sequences covering limited sections of urban, suburban, or highway routes.
This design exposes models to varied scene layouts, traffic patterns, and object distributions, making the datasets highly effective for training and benchmarking.
While these collections provide substantial diversity across many short sequences, they offer only limited long-term temporal continuity and do not capture extended, uninterrupted routes.
Existing autonomous driving datasets often lack high-fidelity digital twins of their recorded environments, creating a significant infrastructure gap.
While generic simulation platforms like CARLA~\cite{CARLA2017Dosovitskiy} are widely used for evaluation~\cite{RobustnessCARLA2023Thirugnana}, their synthetic worlds are not 1-to-1 replicas of real-world routes.
This fundamental lack of geometric and semantic fidelity makes it impossible to conduct robust real-to-sim validation, as scenarios cannot be faithfully transferred.
Only a few publicly available datasets attempt to bridge this gap.
The TWICE dataset~\cite{TWICE2023NovickiNeto} provides a digital twin of a controlled test track, but its coverage is restricted to short, predefined scenarios rather than continuous real-world routes.
CitySim~\cite{CitySim2024Zheng} offers drone-based vehicle trajectory data and 3D maps of the recording sites, yet it lacks sensor data from the ego-vehicle perspective, which is essential for autonomous driving research.
UrbanIng-V2X~\cite{Urbaningv2x2025Roessle} and OPV2V~\cite{OPV2V2022Xu} both feature a 3D digital twin of a small, multi-intersection area for cooperative perception studies.
While these efforts demonstrate the value of digital twins, they are geographically limited to compact urban regions, scenario-specific, and often lack multi-sensor ego-vehicle recordings.
As a result, none provide continuous, route-level real-to-sim mapping suitable for large-scale perception research.
DrivIng addresses these limitations by offering three continuous sequences of an approximately \SI{18}{\kilo\meter} driving route, spanning urban, suburban, and highway segments under day, dusk, and night conditions.
Paired with a geo-referenced digital twin, the dataset enables simulation-driven research, supporting scenario replay, controlled extensions, and systematic evaluation under realistic conditions.
Table~\ref{tbl:map_comparison} shows the key insights of existing datasets that also provide an additional digital twin.

\begin{table}
    \caption{Comparison of driving datasets featuring digital twins.
    Abbreviations: Closed track (C), Urban (U), Highway (H), On-Board Sensor unit (OBS), Road-Side Unit (RSU).}
    \centering
    \setlength{\tabcolsep}{3.7pt}
    \begin{tabular}{lccccc}
        \hline
        Attr. & TWICE & CitySim & OPV2V & UrbanIng- & \textbf{DrivIng} \\
              & \multicolumn{1}{c}{\cite{TWICE2023NovickiNeto}} 
              & \multicolumn{1}{c}{\cite{CitySim2024Zheng}} 
              & \multicolumn{1}{c}{\cite{OPV2V2022Xu}} 
              & \multicolumn{1}{c}{V2X~\cite{Urbaningv2x2025Roessle}} 
              & \multicolumn{1}{c}{(ours)} \\ 
        \hline
        \multicolumn{6}{l}{\textit{--- Core ---}} \\
        Scenes & C & U \& H & U & U & U \& H \\
        Perspective & OBS & Drone & OBS, RSU & OBS, RSU & OBS \\
        \multicolumn{6}{l}{\textit{--- Scale ---}} \\
        Total Area & — & $4~\mathrm{km^2}$  & $4~\mathrm{km^2}$  & $0.64~\mathrm{km^2}$ & $\mathbf{24~\mathbf{km^2}}$ \\
        Drivable Track & — & — & — & \(\sim\!2~\mathrm{km}\) & $\mathbf{\sim\! 18~km}$ \\
        \# Assets & — & — & — & $\sim\!1\,\mathrm{k}$ & \textbf{$>31\,\mathrm{k}$} \\
        \multicolumn{6}{l}{\textit{--- Fidelity ---}} \\
        Ann. Geo-ref. & \ding{53} & \ding{53} & \ding{51} & \ding{51} & \ding{51} \\
        Sign Geo-ref. & \ding{53} & \ding{53} & \ding{53} & \ding{51} & \ding{51} \\
        \hline
    \end{tabular}
    \label{tbl:map_comparison}
\end{table}

\section{Dataset} 
We provide a detailed description of the vehicle sensor suite, including track information, the annotation process, and the construction of the digital twin.
Sensor calibration and synchronization were performed following the procedures described in UrbanIng-V2X~\cite{Urbaningv2x2025Roessle}.

\subsection{Sensor Setup}
\begin{figure}
    \centering
    \includegraphics[width=0.465\textwidth]{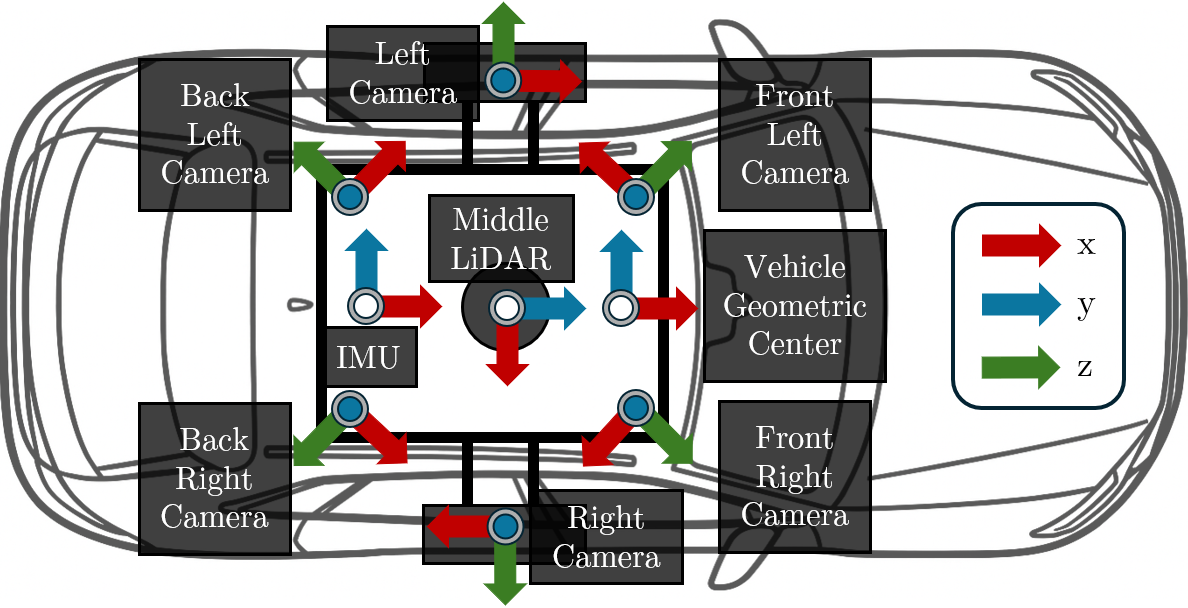}
    \caption{Full sensor setup and coordinate frame of the vehicle.}
    \label{fig:vehicle_sensors}
\end{figure}

Data was collected using an Audi Q8 e-tron equipped with 6 RGB cameras, 1 LiDAR, and 1 GPS/IMU module.  
The cameras are arranged to provide full \SI{360}{\degree} coverage, as illustrated in Figure~\ref{fig:vehicle_sensors}.  
The specifications of each sensor are as follows:
\begin{itemize}
    \item \textbf{RGB Cameras (6×)}: GSML2 SG2-AR0233C-5200-G2A, 20 \gls{fps}, 1920 × 1080 resolution; \SI{60}{\degree} horizontal \gls{fov} (4 cameras), \SI{100}{\degree} horizontal \gls{fov} (2 cameras)
    \item \textbf{LiDAR (1×)}: Robosense Ruby Plus, 20 \gls{fps}, 128 rays, \SI{360}{\degree} horizontal \gls{fov}, \SI{-25}{\degree} to \SI{15}{\degree} vertical \gls{fov}, up to \SI{240}{\meter} range at $\geq$ \SI{10}{\percent} reflectivity
    \item \textbf{GPS/IMU (1×)}: Genesys ADMA Pro+, 100 \gls{fps}, RTK correction, \SI{1}{\centi\meter} precise positioning
\end{itemize}

\subsection{Track Information}
\begin{figure}[b]
    \centering
    \subfloat[Dusk]{%
        \includegraphics[width=0.48\linewidth]{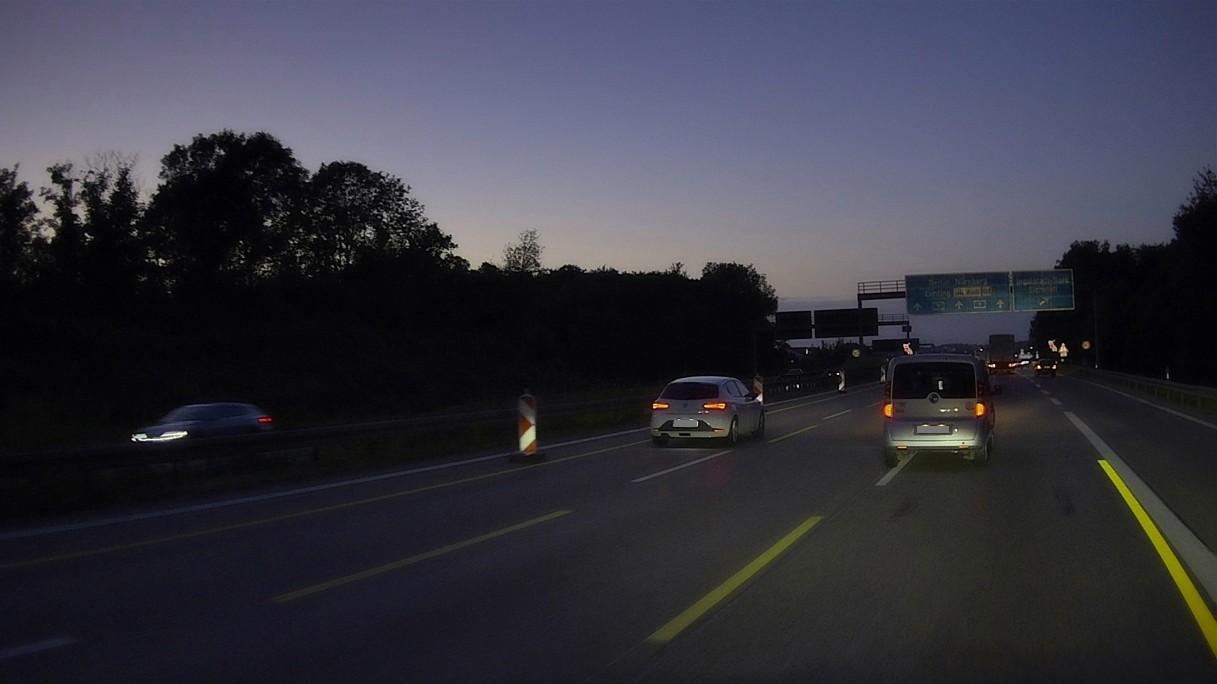}%
        \label{fig:illumdusk}
    }
    \hfill
    \subfloat[Night]{%
        \includegraphics[width=0.48\linewidth]{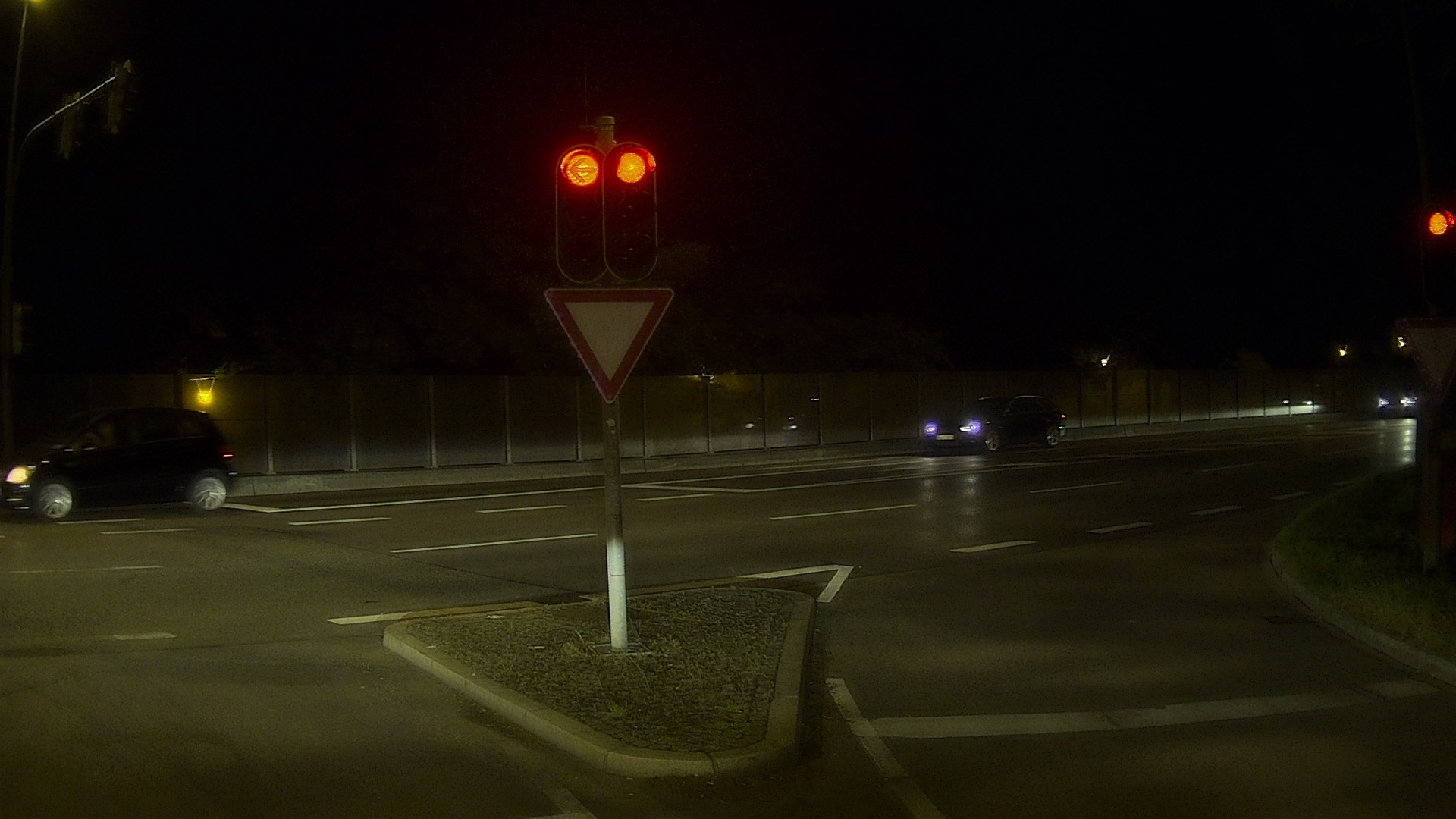}%
        \label{fig:illumnight}
    }
    \caption{Comparison of real-world Dusk and Night illumination.}
    \label{fig:illumination_comparison}
\end{figure}

DrivIng covers an approximately \SI{18}{\kilo\meter} real-world route, comprising over 63k annotated frames, which correspond to about 378k RGB images and 63k LiDAR frames.
Since certain segments of the route are traversed twice in opposite directions, the unique track length amounts to approximately \SI{16}{\kilo\meter}.
The dataset was collected along a track covering highways, suburban streets, urban roads, and several construction zones, with three continuous and uninterrupted sequences recorded under Day, Dusk, and Night lighting conditions.
The Day sequence comprises \num{23092} frames (approx. \SI{38.5}{\minute}), the Dusk sequence comprises \num{20246} frames (approx. \SI{33.7}{\minute}), and the Night sequence comprises \num{19705} frames (approx. \SI{32.8}{\minute}).
The sequences capture a diverse range of road types, varying traffic densities, and representative driving scenarios, including lane changes, merges, and pedestrian crossings.
Figure~\ref{fig:teaser} depicts the complete driven trajectory and the corresponding recording location of a timestamp of the Day sequence.
Figure~\ref{fig:illumination_comparison} additionally shows the illumination of a Dusk and Night frame for better comparison.

\subsection{Annotation Process}
All objects were annotated in the LiDAR point cloud at 10~Hz, including 3D bounding boxes with spatial coordinates ($x, y, z$), yaw orientation, and unique tracking IDs.
Annotation was performed by human annotators, and the quality of the labels was verified through multiple rounds of visual inspection of both the LiDAR point clouds and the corresponding images by independent reviewers to ensure accuracy.
Objects are divided into 12 classes, most of which include additional object-specific attributes (Table~\ref{tbl:attributes}).
To preserve privacy, all visible faces and license plates in the RGB images were anonymized using Gaussian blurring.

\begin{table}[h!]
    \caption{Attribute types and corresponding object classes.}
    \centering
    \begin{tabular}{ll}
        \hline
        \textbf{Attributes} & \textbf{Categories} \\
        \hline
        adult, child & Cyclist, E-Scooter \\
        emergency, regular, public transport & Car, Van, Bus, Truck \\
        car trailer, truck trailer, cyclist trailer & Trailer \\
        standing, walking, sitting & Pedestrian \\
        standing, riding, pushing & Cyclist, E-Scooter \\
        bendy, rigid & Bus, Truck \\
        None & Motorcycle, Animal, Other  \\
         & OtherPedestrian \\
        \hline
    \end{tabular}
    \label{tbl:attributes}
\end{table}

\subsection{Statistics}
\begin{figure*}[ht]
    \centering
    \includegraphics[width=\linewidth]{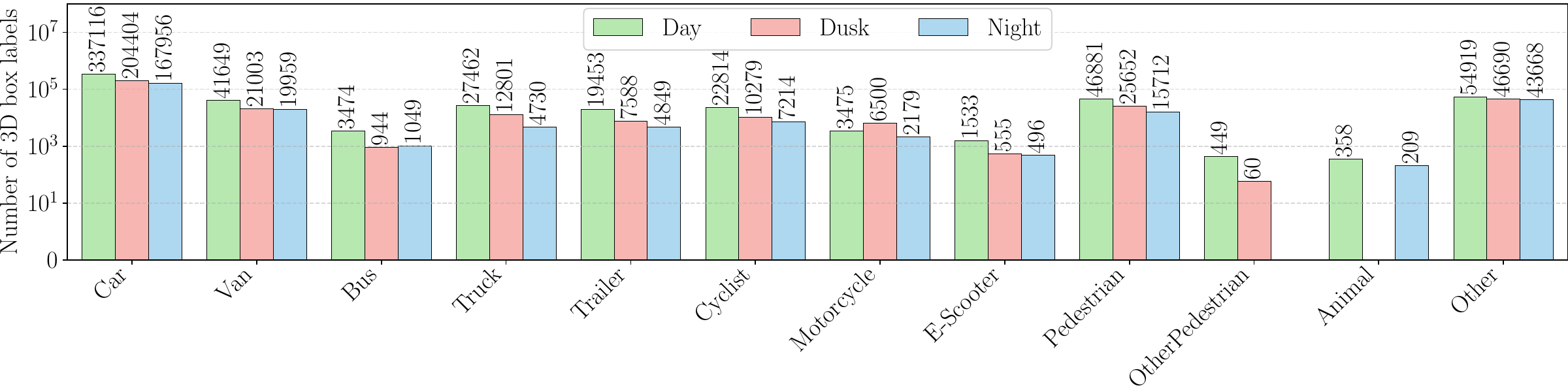}
    \caption{Distribution of all 12 object classes in the dataset, measured by the number of annotated 3D bounding boxes. Cars are the most frequently annotated class, whereas Animals and OtherPedestrian appear least often. Overall, the relative distributions of object classes are consistent across the three recorded sequences.}
    \label{fig:object_class_distributions}
\end{figure*}

\begin{figure*}[ht]
    \centering
    \subfloat[Number of 3D boxes per frame.]{%
        \includegraphics[width=0.32\textwidth]{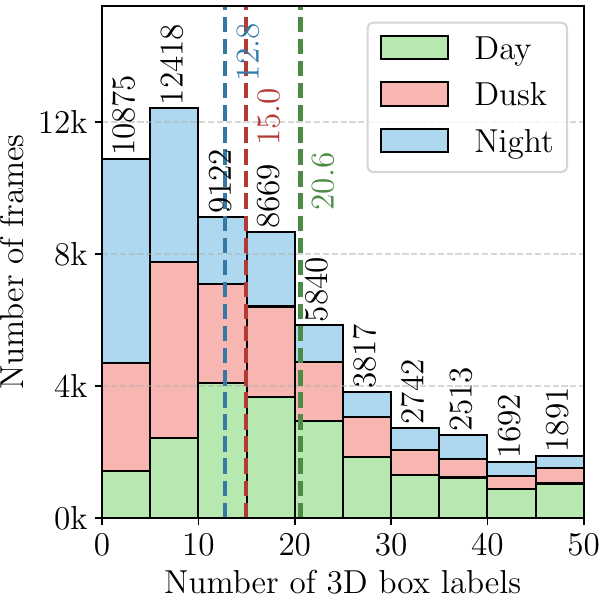}%
        \label{subfig:number3dboxlabels}
    }
    \hfill
    \subfloat[Object rotation distribution.]{%
        \includegraphics[width=0.32\textwidth]{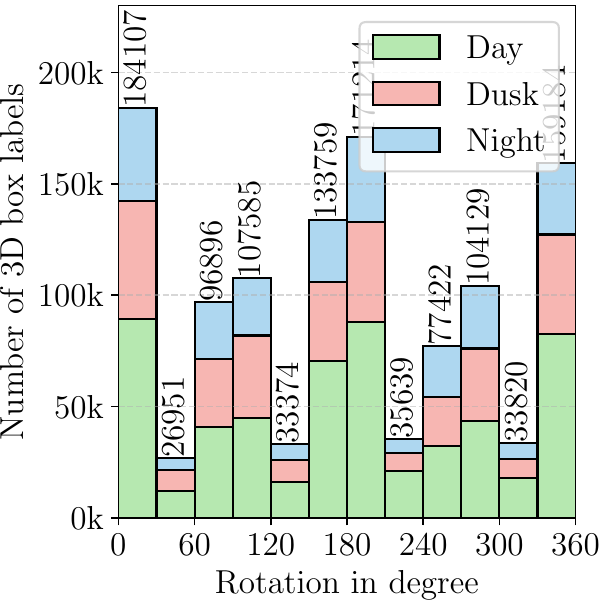}%
        \label{subfig:rotations}
    }
    \hfill
    \subfloat[Number of 3D boxes across distances.]{%
        \includegraphics[width=0.32\textwidth]{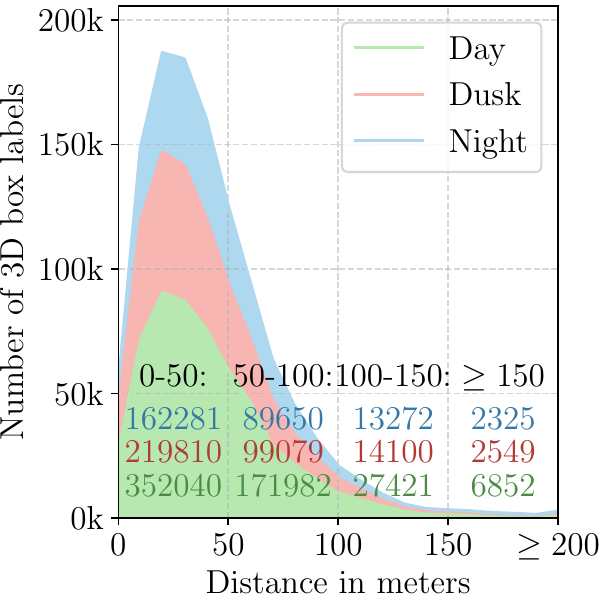}%
        \label{subfig:3dboxesdistance}
    }
    \caption{Visualization (a) illustrates the number of annotated 3D bounding boxes per frame across all sequences. Among the different daytimes, Day contains the highest average number of objects per frame, while Night contains the fewest, yet all sequences include numerous frames with more than 50 objects. Visualization (b) presents the distribution of object orientations relative to the ego vehicle, showing that DrivIng includes a substantial number of objects observed from non-typical traffic angles. Visualization (c) shows the distance distribution of all annotations, with most objects located within \SI{100}{\meter}, while still including a substantial number of objects at longer ranges beyond \SI{100}{\meter}.}
    \label{fig:statistic_figure}
\end{figure*}

\begin{figure*}
    \centering
    \subfloat[Average track length per object class.]{%
        \includegraphics[width=0.49\textwidth]{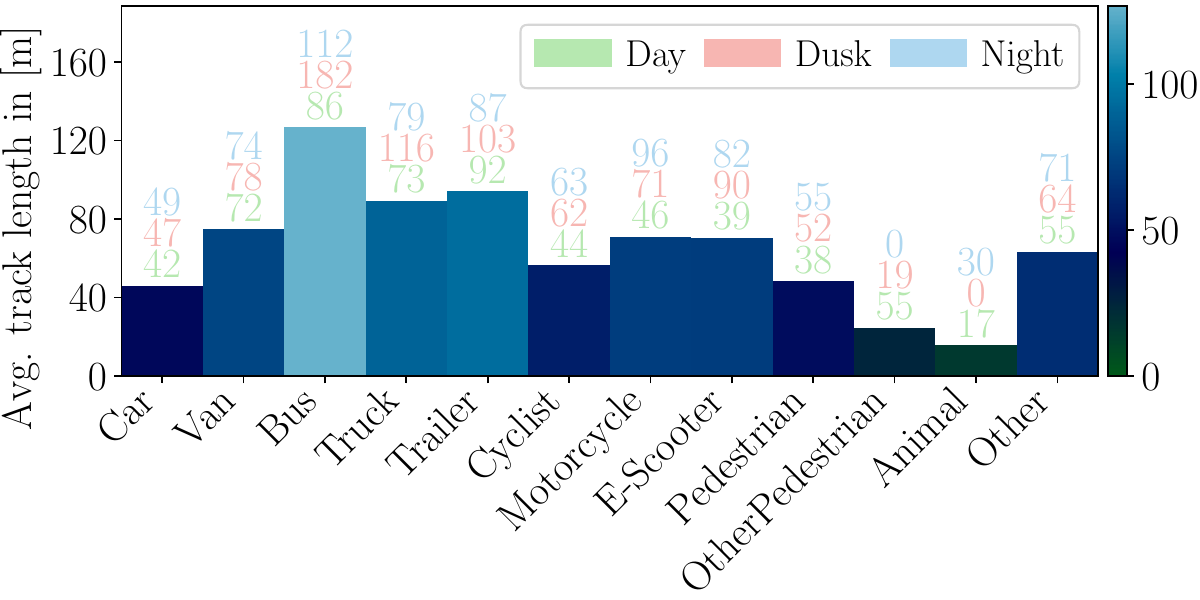}%
        \label{subfig:avgtracklength}
    }
    \hfill
    \subfloat[Average points in 3D boxes per object class.]{%
        \includegraphics[width=0.49\textwidth]{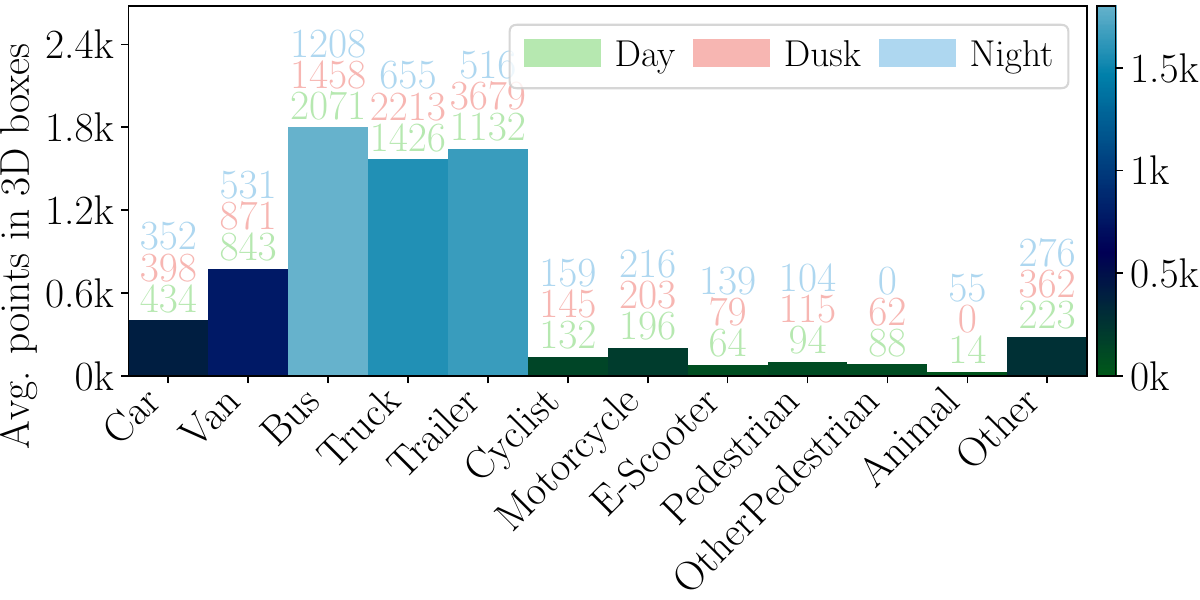}%
        \label{subfig:avgpointsin3dbox}
    }
    \caption{The left visualization (a) shows the average track length, in meters, for each object class. Track lengths are measured per continuous observation segment: if a track ID disappears and later reappears, the later segment is treated as a new distance measurement even when the ID is the same. The right visualization (b) shows the average number of LiDAR points within the 3D bounding boxes for each object class. Across all sequences, larger object classes consistently contain more LiDAR points.}
    \label{fig:avgtrackandpoints}
\end{figure*}

The distribution of all object classes in the DrivIng dataset is shown in Figure~\ref{fig:object_class_distributions}.
In total, the dataset comprises approximately 1.2 million annotated objects, distributed across the three sequences as follows: about 560k in Day, 336k in Dusk, and 268k in Night.
In addition to frequently occurring classes such as \textit{Car}, \textit{Van}, \textit{Pedestrian}, and \textit{Others}, DrivIng also contains annotations for rarer classes like \textit{Animal} and \textit{OtherPedestrian}.
The class \textit{Others} primarily includes construction barriers, cones, and other relevant traffic objects, whereas \textit{OtherPedestrian} refers to pedestrians using supporting devices such as wheelchairs or electric mobility aids. 
Overall, the relative distributions of object classes are consistent across the Day, Dusk, and Night sequences.
The distribution of annotated 3D bounding boxes per frame across the dataset sequences is presented in Figure~\ref{subfig:number3dboxlabels}.
As shown, the Dusk and Night sequences generally contain fewer objects per frame compared to the Day sequence.
This difference primarily reflects typical urban activity patterns, with higher traffic density and pedestrian presence during midday.
On average, the Night sequence contains 12.8 objects per frame, the Dusk sequence 15.0, and the Day sequence 20.6 objects per frame.
Figure~\ref{subfig:rotations} presents the distribution of object orientations, grouped into \SI{30}{degree} bins relative to the ego vehicle.
Most objects, particularly vehicles, are oriented along the primary cardinal directions (north, east, south, and west).
Nevertheless, the dataset also includes a substantial number of objects observed at non-typical traffic angles, reflecting the diversity of real-world traffic patterns.
The distribution of annotated objects with respect to their distance from the ego vehicle is presented in Figure~\ref{subfig:3dboxesdistance}. 
The majority of objects in DrivIng are located within the first \SI{50}{\meter}, accounting for roughly 60\% of all annotations across all sequences. 
Approximately 90\% of the objects lie within a \SI{100}{\meter} range, while the dataset still contains tens of thousands of annotations beyond \SI{100}{\meter}, highlighting its strong coverage of long-range perception scenarios. 
The uninterrupted average track length per object class is shown in Figure~\ref{subfig:avgtracklength}.
Here, a unique track ID is restarted whenever the corresponding object was occluded or unrecognizable for at least one timestamp.
The class \textit{OtherPedestrian} contains no annotations in the Night sequence, and \textit{Animal} has no annotations in the Dusk sequence.
The largest object classes, namely \textit{Bus}, \textit{Truck}, and \textit{Trailer}, as reported in Table~\ref{tbl:object_stats}, exhibit the longest average track lengths.
This trend is consistent with Figure~\ref{subfig:avgpointsin3dbox}, which shows that these same classes also contain the highest average number of LiDAR points within their 3D bounding boxes.
Table~\ref{tbl:object_stats} further provides the number of unique Track IDs across the entire dataset, along with the mean and standard deviation of object-specific lengths, widths, and heights.
Among all object categories, \textit{Bus}, \textit{Truck}, and \textit{Trailer} are the largest, whereas \textit{Animal} and \textit{Pedestrian} (including \textit{OtherPedestrian}) represent the smallest.

\begin{table}[ht]
    \centering
    \setlength{\tabcolsep}{5pt} 
    \caption{Statistics of object dimensions across classes, showing mean and standard deviation (in meters). ''OtherPed.`` refers to object class OtherPedestrian. The highest values for Track IDs and Mean are highlighted in bold, whereas the lowest values are underlined.}
    \begin{tabular}{lccccccc}
        \hline
        \multirow{2}{*}{Obj. Class} & 
        \multirow{2}{*}{Track IDs} & 
        \multicolumn{2}{c}{Length} & 
        \multicolumn{2}{c}{Width} & 
        \multicolumn{2}{c}{Height} \\
        \cline{3-4} \cline{5-6} \cline{7-8}
         &  & Mean & Std & Mean & Std & Mean & Std \\
        \hline
        Car & \textbf{8502} & 4.33 & 0.79 & 1.88 & 0.31 & 1.65 & 0.34 \\
        Van & 640 & 5.45 & 0.95 & 2.22 & 0.37 & 2.31 & 0.44 \\
        Bus & 41 & \textbf{10.46} & 1.82 & \textbf{3.12} & 0.44 & 3.35 & 0.29 \\
        Truck & 343 & 8.07 & 3.61 & 3.09 & 0.87 & \textbf{3.45} & 0.90 \\
        Trailer & 261 & 9.40 & 5.12 & 2.85 & 0.97 & 3.17 & 1.25 \\
        Cyclist & 299 & 1.96 & 0.42 & 0.88 & 0.25 & 1.74 & 0.39 \\
        Motorcycle & 74 & 2.15 & 0.43 & 0.93 & 0.22 & 1.57 & 0.21 \\
        E-Scooter & 25 & 1.21 & 0.37 & 0.76 & 0.20 & 1.78 & 0.39 \\
        Pedestrian & 758 & \underline{0.75} & 0.21 & 0.70 & 0.17 & 1.66 & 0.25 \\
        OtherPed. & \underline{5} & 1.00 & 0.23 & 0.68 & 0.11 & 1.16 & 0.17 \\
        Animal & 6 & 0.98 & 0.20 & \underline{0.49} & 0.15 & \underline{0.62} & 0.09 \\
        Other & 1094 & 1.26 & 2.36 & 2.38 & 3.79 & 1.42 & 0.42 \\
        \hline
    \end{tabular}
    \label{tbl:object_stats}
\end{table}

\subsection{The Digital Twin}\label{subsec:digitaltwin}
\begin{figure}
    \centering
    \includegraphics[width=0.485\textwidth]{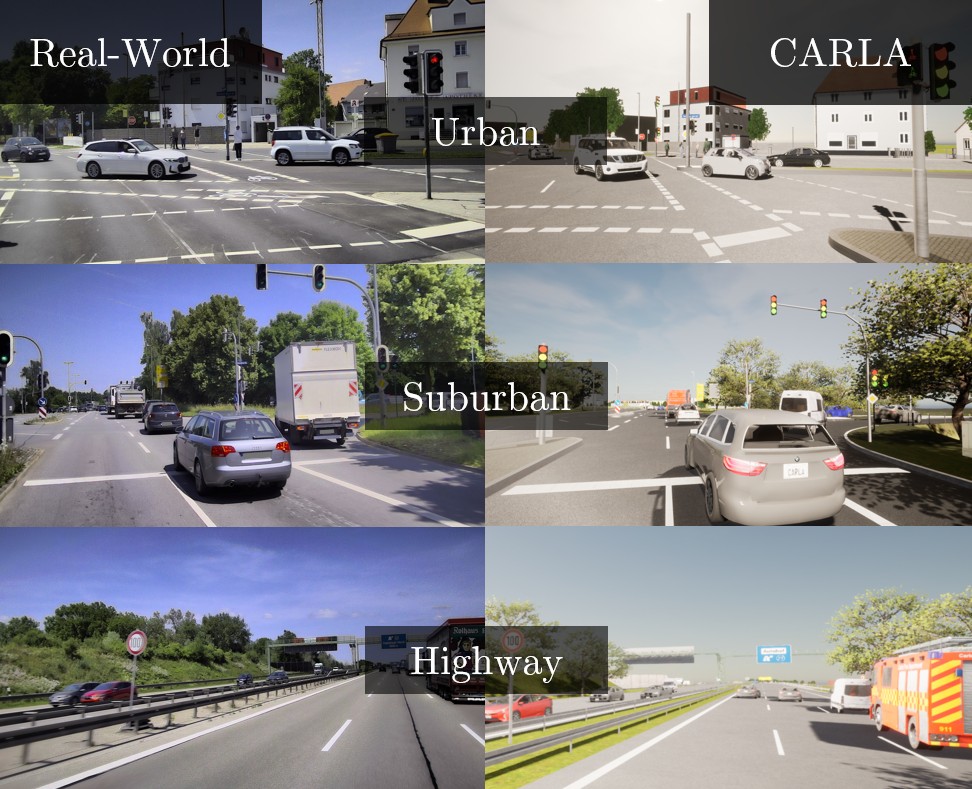}
    \caption{Real-world and CARLA digital twin views at matched locations. All elements of the digital twin, including road topology, buildings, and scene assets, are fully geo-referenced to their real-world counterparts.}
    \label{fig:carla_realworld_comparison}
\end{figure}

A core contribution of our work is the development of a large-scale digital twin of the full data recording route in CARLA~\cite{CARLA2017Dosovitskiy}, which provides a comprehensive, geo-referenced reconstruction of the $6 \times 4$ km$^2$ data collection area.
The digital twin is anchored by a detailed HD map that links the simulation to precise global coordinates.
Built from independent geo-referenced recordings of the real-world route, the 3D environment was enriched with over 1.2k hand-crafted buildings, more than 10k traffic signs, and over 20k additional environmental objects, ensuring a high-fidelity real-to-sim correspondence.
Covering the full \SI{18}{\kilo\meter} real-world drivable route, the resulting 3D map provides accurate geo-referencing and a reliable foundation for large-scale simulation experiments.
As illustrated in Figure~\ref{fig:carla_realworld_comparison}, the real-world data can be seamlessly integrated into the simulation, producing exact correspondences for both dynamic and static objects.
Using only timestamps and 3D annotations, the digital twin reconstructs the scene by placing surrogate vehicle models at the same global coordinates as their real-world counterparts.

The primary strength of our digital twin lies in its ability to reconstruct real-world scenarios in two complementary modes: high-fidelity kinematic replay (Algorithm~\ref{alg:kinematic}) and live interactive re-simulation (Algorithm~\ref{alg:interactive}).
In kinematic replay mode, all agent trajectories are replayed exactly as recorded, forcing each agent to follow its precise spatial and temporal path from the real-world dataset.
This capability is essential for sensor-level validation, allowing for a direct comparison between real-world sensor data and its simulated counterpart.
For this mode, we select a surrogate model by matching the real-world agent's object class and its dimension to the geometrically closest vehicle in our library.
Our approach resets the scene at every frame using ground-truth agent positions and orientations from the dataset.
Agents are placed directly at their recorded poses rather than propagated over time, completely bypassing CARLA’s physics engine.
Consequently, any positional discrepancy arises solely from CARLA’s transform precision and reflects simulator limitations rather than trajectory reconstruction error.

In interactive re-simulation mode, the real-world data can be used to extract an agent’s initial state and its recorded trajectory, which then serves as a global reference path.
CARLA’s built-in AI pilot may be assigned to follow this path while autonomously managing local interactions, such as yielding during lane changes or responding to surrounding traffic.
This enables the creation of live, interactive, physics-based simulations suitable for evaluating planning and control systems.

\begin{algorithm}[t]
    \caption{Kinematic Replay Mode}
    \label{alg:kinematic}
    \begin{algorithmic}[1]
        \REQUIRE Real-world dataset $\mathcal{D}$
        \STATE Set synchronous mode ($\Delta t = 100$ ms)
        \FOR{each frame $f$ in $\mathcal{D}$}
            \STATE Clear all actors
            \STATE Place ego vehicle at $f.\text{gps\_pose}$
            \FOR{each agent $a$ in $f.\text{annotations}$}
                \STATE Select surrogate model matching $a.\text{class}$ and $a.\text{dimensions}$
                \STATE Spawn agent at $a.\text{centroid}$ with $a.\text{orientation}$
            \ENDFOR
            \STATE Record sensor data
            \STATE Advance simulation by one frame
        \ENDFOR
    \end{algorithmic}
\end{algorithm}

\begin{algorithm}[t]
    \caption{Interactive Re-simulation Mode}
    \label{alg:interactive}
    \begin{algorithmic}[1]
        \REQUIRE Real-world dataset $\mathcal{D}$, test policy $\pi$
        \STATE Set synchronous mode ($\Delta t = 100$ ms)
        \STATE Initialize scenario using frame 0 of $\mathcal{D}$
        \FOR{each agent $a$}
            \STATE Enable CARLA autopilot
            \STATE Set reference path to $a$'s recorded trajectory
        \ENDFOR
        \WHILE{scenario active}
            \STATE Apply control command from $\pi$ to ego vehicle
            \STATE Advance simulation by one frame
            \STATE Record states and interactions
        \ENDWHILE
        \STATE \textbf{Note:} Initial state precision matches Algorithm~\ref{alg:kinematic}
    \end{algorithmic}
\end{algorithm}

A remaining limitation in both modes is the visual fidelity of agents, which is constrained by the finite set of vehicle models provided by the simulator.

\section{Tasks}
The DrivIng dataset provides rich 3D annotations that enable multiple perception tasks, such as object detection, tracking, trajectory prediction, and localization.
In this work, however, we focus on 3D object detection.
To ensure compatibility with established benchmarks, we transform our dataset into the nuScenes format~\cite {NuScenes2020Caesar} and leverage the MMDetection3D Pipeline~\cite {MMDet3d2020MMDet3dcontributors} for both training and evaluation.
During this conversion, several of the dataset’s original categories are merged to align with the predefined nuScenes object classes.
Out of the dataset’s 12 annotated categories, this mapping yields a final set of 9 nuScenes classes, as shown in Table~\ref{tbl:org_to_nuscenes_mapping}.
We also excluded the class Animal due to its underrepresentation, resulting in a total of 8 classes for the 3D object detection task.
\begin{table}
    \caption{Mapping from original dataset object classes to nuScenes object classes.}
    \centering
    \begin{tabular}{ll}
        \hline
        \textbf{Original Categories} & \textbf{nuScenes Class} \\
        \hline
        Car, Van & vehicle.car \\
        Truck & vehicle.truck\\
        Bus & vehicle.bus \\
        Trailer & vehicle.trailer \\
        E-Scooter, Cyclist & vehicle.bicycle \\
        Motorcycle & vehicle.motorcycle \\
        Pedestrian, OtherPedestrian & human.pedestrian.adult \\
        Animal & animal \\
        Other & movable\_object.barrier \\
        \hline
    \end{tabular}
    \label{tbl:org_to_nuscenes_mapping}
\end{table}
For benchmarking, we follow the nuScenes evaluation protocol, reporting Average Translation Error (ATE), Average Scale Error (ASE), Average Orientation Error (AOE), and Average Velocity Error (AVE).
We further include the nuScenes distance-aware mAP, which computes precision across object categories and four predefined distance thresholds (\SI{0.5}{\meter}, \SI{1}{\meter}, \SI{2}{\meter}, \SI{4}{\meter}), as well as the nuScenes Detection Score (NDS).
Since the object attributes in our dataset differ substantially from those in nuScenes, we exclude the Average Attribute Error (AAE) from our evaluation.

\section{Experiments}
We benchmark 3D object detection performance using two state-of-the-art models: PETR~\cite{Liu2022PETR} for camera-only input and CenterPoint~\cite{Yin2021CenterPoint} for LiDAR-only input.
PETR leverages a pre-trained FCOS3D~\cite{Wang2021FCOS3D} (V-99-eSE) backbone with input images resized to 384 × 960 pixels and is trained for 75 epochs.
CenterPoint operates on a single LiDAR sweep projected onto a \SI{100}{\meter} $\times$ \SI{100}{\meter} BEV grid with a voxel size of (0.1, 0.1, 0.2) meters and is trained for 25 epochs.
For consistency, the detection heads of both models are adapted to our eight-class setting (excluding the animal category).
The evaluation ranges are set to standard nuScenes limits of $[\SI{-54}{\meter}, \SI{54}{\meter}]$ along both x and y axes.
The full experiment setup can be found in our GitHub repository.

\subsection{Dataset Configuration}
Because the recordings were collected at different times of day, we keep the day, dusk, and night sequences as they are for training and testing.
Specifically, for each sequence, we create a training, validation, and test set.
We split each full sequence into 50 sub-sequences.
Within each sub-sequence, 80\% were used for training, 10\% for validation, and the remaining 10\% for testing.
This split ensures that every partition of the sequence includes coverage of all environment types, including highway, suburban, and urban scenes.
We train and evaluate PETR~\cite{Liu2022PETR} and CenterPoint~\cite{Yin2021CenterPoint} on each sequence individually.
All models are trained on six NVIDIA L40S GPUs paired with an Intel\textsuperscript{\textregistered} Xeon\textsuperscript{\textregistered} Platinum 8480+ processor with 224 cores.
For reference, training CenterPoint~\cite{Yin2021CenterPoint} for 25 epochs on sequence Day with a batch size of 4 on each GPU requires approximately 18 hours.

\subsection{Benchmark results}
Table \ref{tbl:nuscenes_results} reports nuScenes evaluation results across the three sequences.
PETR, the camera-based model, consistently achieves lower mAP and NDS than the LiDAR-based CenterPoint, with CenterPoint reaching at least twice the mAP of PETR and nearly three times the value at night.
PETR shows lower AVE, indicating better velocity estimates.
Both models show performance degradation from day to night, with CenterPoint being affected by the fewer nearby objects in the sequences and PETR being further impacted by low illumination and weaker geometric cues.
The per-class AP in Table~\ref{tbl:ap_results} shows that CenterPoint outperforms PETR on small classes, such as Bicycle and Pedestrian.
PETR also struggles with larger objects, such as trailers, due to higher localization and orientation errors, sparse or partially visible structures, and high appearance variability, which amplify misalignment for these long articulated objects.

\begin{table}
    \caption{NuScenes evaluation metrics on PETR and CenterPoint for all three sequences. PETR uses camera-only input. CenterPoint uses LiDAR-only input.
    }
    \centering
    \setlength{\tabcolsep}{4.15pt} 
    \begin{tabular}{llcccccc}
        \hline
        \textbf{Test Set} & \textbf{Model} & \textbf{ATE} & \textbf{ASE} & \textbf{AOE} & \textbf{AVE} & \textbf{NDS} & \textbf{mAP} \\
        \hline
        \multirow{2}{*}{Day} 
         & PETR~\cite{Liu2022PETR}                  & 0.70 & 0.19 & 0.27 & 1.54 & 40.4 & 35.9 \\
         & CenterPoint~\cite{Yin2021CenterPoint}    & 0.20 & 0.16 & 0.16 & 3.72 & 71.7 & 79.3 \\
        \hline
        \multirow{2}{*}{Dusk} 
         & PETR~\cite{Liu2022PETR}                  & 0.73 & 0.28 & 0.36 & 2.46 & 38.0 & 35.7 \\
         & CenterPoint~\cite{Yin2021CenterPoint}    & 0.16 & 0.17 & 0.27 & 3.22 & 66.9 & 72.5 \\
        \hline
        \multirow{2}{*}{Night} 
         & PETR~\cite{Liu2022PETR}                  & 0.99 & 0.28 & 0.41 & 2.16 & 27.2 & 22.4 \\
         & CenterPoint~\cite{Yin2021CenterPoint}    & 0.23 & 0.17 & 0.22 & 3.87 & 63.1 & 66.1 \\
        \hline
    \end{tabular}
    \label{tbl:nuscenes_results}
\end{table}

\begin{table*}
    \caption{Per-class AP averaged across AP@(\SI{0.5}{\meter}, \SI{1}{\meter}, \SI{2}{\meter}, \SI{4}{\meter}) evaluated on PETR and CenterPoint and all three sequences. Abbreviations: trk = truck, trl = trailer, bic = bicycle, ped = pedestrian, mot = motorcycle, bar = barrier. PETR uses camera-only input. CenterPoint uses LiDAR-only input.
    }
    \centering
    \begin{tabular}{llcccccccccc}
        \hline
        \textbf{Test Set} & \textbf{Model} & $\mathbf{AP_\text{car}}$ & $\mathbf{AP_\text{trk}}$ & $\mathbf{AP_\text{bus}}$ & $\mathbf{AP_\text{trl}}$ & $\mathbf{AP_\text{bic}}$ & $\mathbf{AP_\text{mot}}$ & $\mathbf{AP_\text{ped}}$ & $\mathbf{AP_\text{bar}}$ & $\mathbf{mAP}$\\
        \hline
        \multirow{2}{*}{Day} 
         & PETR~\cite{Liu2022PETR}                  & 48.8 & 40.1 & 28.3 & 10.6 & 27.1 & 60.8 & 27.8 & 43.9 & 35.9 \\
         & CenterPoint~\cite{Yin2021CenterPoint}    & 90.8 & 79.9 & 58.6 & 64.6 & 88.1 & 94.9 & 80.2 & 77.7 & 79.3 \\
        \hline
        \multirow{2}{*}{Dusk} 
         & PETR~\cite{Liu2022PETR}                  & 42.5 & 52.2 & 18.7 & 17.8 & 28.3 & 63.2 & 16.3 & 46.6 & 35.7 \\
         & CenterPoint~\cite{Yin2021CenterPoint}    & 88.0 & 71.9 & 47.9 & 47.3 & 80.5 & 87.7 & 78.7 & 78.0 & 72.5 \\
        \hline
        \multirow{2}{*}{Night} 
         & PETR~\cite{Liu2022PETR}                  & 35.2 & 11.9 & 27.5 & 0.0 & 27.5 & 29.6 & 16.0 & 31.6 & 22.4 \\
         & CenterPoint~\cite{Yin2021CenterPoint}    & 89.2 & 68.2 & 52.3 & 22.0 & 82.5 & 83.2 & 71.1 & 60.3 & 66.1\\
        \hline
    \end{tabular}
    \label{tbl:ap_results}
\end{table*}

\section{Conclusion}
In this work, we introduce \textbf{DrivIng}, a large-scale, multimodal dataset designed to advance research in autonomous driving.
The dataset includes three continuous, full-length sequences spanning roughly \SI{18}{\kilo\meter} under varied lighting conditions.
It covers highways, suburban and urban roads, and multiple construction zones, offering diverse driving scenarios.
DrivIng provides comprehensive $360^{\circ}$ perception with 6 RGB cameras, one LiDAR sensor, and a high-precision ADMA system for accurate geo-referencing.
All sensors are carefully calibrated both extrinsically and intrinsically, and temporally synchronized to ensure precise multimodal alignment.
We provide over 63k frames at 10~Hz with a total of approximately 1.2 million annotated objects, including 3D bounding boxes with tracking IDs across 12 object classes with additional specific attributes.
In addition to detailed dataset statistics, we provide baseline results for the camera and LiDAR models using \gls{sota} 3D object detectors.
Furthermore, we provide a high-fidelity, geo-referenced digital twin of the entire driving route, which functions as a paired validation testbed capable of precisely reconstructing any real-world event from our dataset.
This resource supports a wide range of future research applications, including sim-to-real transfer, robust evaluation under controlled environmental changes, systematic testing of safety-critical scenarios, deployment of complex multi-agent algorithms for cooperative perception, and improved generalization to real-world conditions.
To promote adoption and reproducibility, we release the complete codebase, development kit, the real-world DrivIng dataset, and its digital twin, along with full integration into MMDetection3D and conversion scripts for both real and simulated data in the nuScenes format.

\section*{Acknowledgments}
This work was partially funded by the Bavarian state government as part of the High Tech Agenda, the iEXODDUS project (GA 101146091), and the Bavarian Academic Forum (BayWISS).

\bibliographystyle{IEEEtran}
\bibliography{literature}

\end{document}